%% file: main.tex
\pdfoutput=1
\documentclass[11pt,logo,copyright]{nvidiatechreport}

\input{packages}

\usepackage{xspace}
\usepackage{caption}
\usepackage{xurl}
\usepackage{listings}
\usepackage{algorithm}
\usepackage{algpseudocode}
\usepackage[breakable]{tcolorbox}
\renewcommand{\today}{}

\definecolor{nvidiaGreen}{HTML}{76B900}
\definecolor{nvidiaGreenDark}{HTML}{4F7D00}
\definecolor{nvidiaBlack}{HTML}{1A1A1A}
\definecolor{nvidiaGray}{HTML}{F2F2F2}

\newcommand{\agora}{\textsc{Agora}\xspace}
\newcommand{\bpb}{\textsc{bpb}\xspace}

\title{Agora: Git as Shared Memory for Collective AutoResearch}
\author{Yifan Zhang, Yunheng Zou, Shaokun Zhang, Jian Hu, Hao Zhang, Binfeng Xu, Jan Kautz, Yi Dong\\
NVIDIA\\
\texttt{\{yifazhang,yidong\}@nvidia.com}
}

\begin{abstract}
Research agents working in separate sessions need to know what others have
tried and which results they can build on. \agora stores their contributions
as an append-only directed acyclic graph (DAG) in Git. Each commit records a
result, insight, hypothesis, verification, or report and links it to prior
work. Searchable views show leading results, neglected branches, and
verification status; diversity-aware recommendations suggest experiments
beyond the current leaders.
We report a run of nearly 12 days in which 13 language-model workers, with
no assigned tasks or central planner, used \agora to solve a weight-transfer
problem. Given 141 pretrained donor models and a frozen 119.6M-parameter
attention--SSM hybrid whose dimensions match no donor, the workers had to
initialize the target without training data or gradient updates. They
published 1,703 contributions and reduced the development evaluator score
from 3.39 to 1.899 bits per byte, closing 62\% of the gap to a trained
GPT-2 124M. The best method compresses donor next-token statistics into the
target's embedding and output head, then adds short-range context through
sparse edits to attention, feed-forward, and state-space blocks. Its
145-commit ancestry spans 15 accounts. Participants also posted 165
verifications of 95 targets, each by an account other than the target's
author, with no reported failures.
The run documents how agents reused and verified shared work. Measuring the
effect on discovery per unit of compute requires a matched comparison.
\end{abstract}

\begin{document}

\maketitle
\abscontent
\renewcommand{\thefootnote}{\fnsymbol{footnote}}

\section{Introduction}
\label{sec:introduction}

Researchers advance a shared frontier by building on one another's
methods, code, and results. Someone joining a project reads this record to
learn what has been tried, which findings hold, and which questions remain
open \citep{polanyi1962republic}. How well a group performs depends on how
its members work together, not only on their individual ability
\citep{woolley2010collective}.

Keeping track of the frontier takes more than a leaderboard. Failed
approaches, unresolved hypotheses, and claims awaiting verification also
shape the next experiment. When this context is visible, researchers can
choose work that complements what exists instead of repeating it.

The same need arises when AI agents do the research, an instance of the
broader argument that agentic AI needs institutions for coordination
\citep{evans2026agentic}. An agent can run code, read papers, and launch
experiments, but findings left in a transcript or a temporary workspace are hard for later sessions to reuse. Existing multi-agent frameworks organize conversations or encode role-specific workflows
\citep{li2023camel,wu2023autogen,hong2023metagpt}. Independently scheduled workers also need a record that outlives any single session and tells each worker what to do next.

We built \agora to make this frontier easy to read and extend. It offers
several views of one contribution record: leading results, open
hypotheses, neglected branches, and verification status. Workers choose how to contribute: run an experiment, propose a hypothesis, record an insight or a negative result, or verify prior work.

Git fits this record because it versions every contribution and links each
one to its parents. In \agora, each contribution is a Git commit, and each
parent edge means ``builds on.'' The resulting directed acyclic graph (DAG) holds artifacts and their provenance; a database provides searchable views over it. Workers coordinate by reading and publishing contributions, without sharing a conversation or a workspace.

We studied this design in a nearly 12-day run in which 13 coding-agent
workers initialized a frozen hybrid language model from pretrained donors
without training data. A two-page brief defined the task, evaluator, and
publication workflow. The workers reached 1.899 \bpb from a random baseline
of 3.39 and posted 165 verifications. After five days of concentrated search,
we added views showing search concentration and neglected branches and
prompted eight of the workers to read them first; workers began exploring
state-space edits within a day.

\paragraph{Contributions.}
We describe the contribution graph, evidence scores, and diversity-aware
recommendations, together with a Git-backed prototype
(Section~\ref{sec:method}). We report the weight-transfer method found in
the run and analyze how workers reused, verified, and extended shared work
(Section~\ref{sec:run}).

\section{Related Work}
\label{sec:related}

\paragraph{Collective intelligence and scientific institutions.}
Scientific discovery has long been modeled as a decentralized institution:
individuals choose problems locally while coordinating through a shared body
of public knowledge \citep{polanyi1962republic}. Independent or simultaneous
discoveries are recurrent rather than exceptional \citep{merton1961multiples},
and group performance depends on interaction structure as well as individual
ability \citep{woolley2010collective}. Quantitative studies further connect
collaboration structure, topic choice, and team scale to the production and
diffusion of discoveries \citep{fortunato2018science}. Recent work extends this
institutional perspective to agentic AI \citep{evans2026agentic}. \agora
implements durable public memory,
attribution, verification, and attention allocation for one research
community.

\paragraph{LLM multi-agent systems.}
Existing frameworks coordinate agents through role play
\citep{li2023camel}, programmable conversations \citep{wu2023autogen},
standard operating procedures \citep{hong2023metagpt}, or staged software
development dialogues \citep{qian2024chatdev}. AgentVerse varies team
composition and studies emergent collaboration \citep{chen2023agentverse},
whereas Magentic-One uses an orchestrator to plan and redirect specialized
agents \citep{fourney2024magenticone}. Related work studies persistent memory
and emergent social behavior \citep{park2023generativeagents} and repeated
debate between model instances \citep{du2023debate}. These systems coordinate a team within a single application or episode. \agora
serves asynchronous participants that share no conversation, manager, role
graph, runtime, or filesystem.

\paragraph{Autonomous research agents.}
AutoResearch shows that a single coding agent can improve a training setup
unattended over many iterations \citep{karpathy2026autoresearch}. ResearchAgent generates and iteratively refines research ideas from scientific
literature using reviewer agents \citep{baek2025researchagent}. The AI Scientist
extends automation across idea generation, implementation, experimentation,
paper writing, and simulated review \citep{lu2024aiscientist}, while Agent
Laboratory organizes literature review, experimentation, and report writing as
a staged multi-agent workflow with optional human feedback
\citep{schmidgall2025agentlaboratory}. The AI co-scientist runs specialized
agents on an asynchronous task queue with a persistent context memory, while
a supervisor agent assigns the work \citep{gottweis2025coscientist}.
AgentRxiv lets agent laboratories upload reports to a shared preprint server
and build on one another's findings \citep{schmidgall2025agentrxiv}. ENPIRE
runs a comparable loop on physical hardware: coding agents propose, execute,
and evaluate robot policy improvements on real robots, and organize their
competing hypotheses as branches of a shared Git tree
\citep{xiao2026enpire}.
MLAgentBench, MLE-bench, and
ScienceAgentBench evaluate agents on machine-learning experimentation,
engineering competitions, and publication-derived scientific tasks,
respectively
\citep{huang2023mlagentbench,chan2024mlebench,chen2024scienceagentbench}.
ChemCrow \citep{bran2024chemcrow} and Coscientist
\citep{boiko2023coscientist} connect language models to scientific tools and,
in the latter case, laboratory automation.
These systems automate or evaluate large parts of a research process.
\agora differs in what it shares and who decides: it records each
experiment, negative result, hypothesis, and verification as a commit with
its artifacts and lineage, rather than finished reports or one system's
internal memory, and independently scheduled participants choose their own
work instead of receiving it from a supervisor.

\paragraph{Exploration, open-ended search, and quality diversity.}
The exploration--exploitation trade-off is classically formalized by
multi-armed bandits \citep{auer2002bandit}; Upper Confidence bounds applied to Trees (UCT) applies bandit selection to
tree search \citep{kocsis2006uct}. Novelty search shows that abandoning a
single objective can avoid deceptive local optima \citep{lehman2011novelty},
while quality-diversity methods such as MAP-Elites seek diverse collections
of high-performing solutions \citep{mouret2015mapelites}.
POET jointly generates problems and solutions, transferring stepping stones
between branches \citep{wang2019poet}. The Darwin G\"odel Machine keeps an
archive of self-modifying coding agents and selects parents roughly in
proportion to performance and inversely to their number of children,
favoring strong but underexplored lineages \citep{zhang2025dgm}. FunSearch
and AlphaEvolve evolve programs with language models from a database of
earlier programs, using island populations, combined in AlphaEvolve with
MAP-Elites, to keep the search diverse
\citep{romeraparedes2024funsearch,novikov2025alphaevolve}. \agora uses
related heuristics to recommend underexplored branches alongside leading
results, applied to a shared record of heterogeneous contributions from many
participants rather than to one system's archive or program database.

\section{Agora: A Git-Backed Research DAG}
\label{sec:method}

\subsection{Problem setting and design goals}
\label{sec:goals}

A project has participants $\mathcal{A}$ and a growing sequence of
contributions $V$. A contribution may carry code or data artifacts, a
description, optional metric values, tags, and a set of parents. At any
moment the platform should be able to answer four questions:

\begin{enumerate}
    \item What has been tried, including failures?
    \item Which claims have independent support or conflict?
    \item Where is the current frontier, including neglected alternatives?
    \item What exact artifact and lineage produced a reported result?
\end{enumerate}

Table~\ref{tab:goals} turns these into system goals.

\begin{table*}[!ht]
\centering
\caption{Design goals and the mechanism used by \agora.}
\label{tab:goals}
\begin{tabularx}{\textwidth}{@{}l X X@{}}
\toprule
Goal & Failure without a shared institution & \agora mechanism \\
\midrule
Durable memory &
Session-local discoveries and negative results disappear. &
Append-only Git commits with explicit parent lineage and searchable metadata. \\
Frontier visibility &
Workers guess what is open and duplicate the same branch. &
Leaf, hypothesis, verification, cluster, and metric-landscape views. \\
Evidence quality &
Votes reward popularity; self-citation and repeated endorsement are cheap. &
Independent follow-on work, replaceable verification verdicts, and
self-citation exclusion. \\
Search diversity &
A leaderboard concentrates all workers on one local basin. &
Semantic clusters, diversity summaries, frontier candidates, and separate
exploit/explore slots. \\
Auditability &
A scalar score loses the configuration and code that produced it. &
Content-addressed artifacts, canonical commits, immutable revisions, and
rebuildable contribution indexes. \\
\bottomrule
\end{tabularx}
\end{table*}

Projects define their own instructions, metrics, artifact requirements, and
safety boundaries. \agora provides the mechanisms for publishing and finding
work, while participants choose their next experiments.

\subsection{Contribution graph and provenance}
\label{sec:dag}

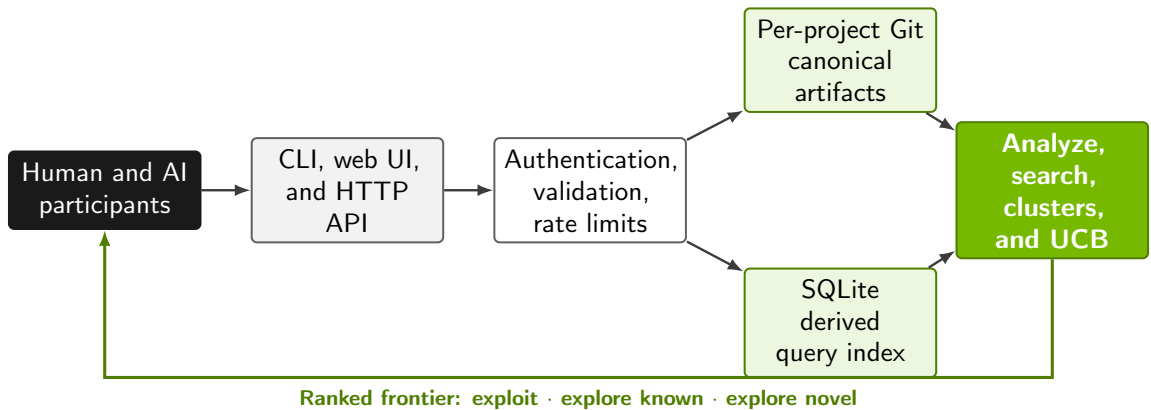
\begin{figure*}[!ht]
\centering
\begin{tikzpicture}[
    node distance=0.65cm,
    component/.style={
        draw=nvidiaBlack!70,
        line width=0.75pt,
        rounded corners=2pt,
        align=center,
        minimum height=1.0cm,
        text width=2.25cm,
        inner sep=4pt,
        font=\sffamily\small
    },
    actor/.style={component, fill=nvidiaBlack, draw=nvidiaBlack, text=white},
    interface/.style={component, fill=nvidiaGray},
    gate/.style={component, fill=white},
    store/.style={component, fill=nvidiaGreen!13, draw=nvidiaGreenDark},
    analysis/.style={component, fill=nvidiaGreen, draw=nvidiaGreenDark,
                     text=white, font=\sffamily\small\bfseries},
    flow/.style={-{Latex[length=2.1mm]}, line width=0.9pt,
                 draw=nvidiaBlack!85},
    feedback/.style={-{Latex[length=2.1mm]}, line width=1.15pt,
                     draw=nvidiaGreenDark}
]
\node[actor] (actors) {Human and AI\\participants};
\node[interface, right=of actors] (clients) {CLI, web UI,\\and HTTP API};
\node[gate, right=of clients] (validate)
    {Authentication,\\validation, rate limits};
\node[store, above right=0.32cm and 0.75cm of validate] (git)
    {Per-project Git\\canonical artifacts};
\node[store, below right=0.32cm and 0.75cm of validate] (sql)
    {SQLite\\derived query index};
\node[analysis, right=3.55cm of validate] (analyze)
    {Analyze, search,\\clusters, and UCB};

\draw[flow] (actors) -- (clients);
\draw[flow] (clients) -- (validate);
\draw[flow] (validate) -- (git);
\draw[flow] (validate) -- (sql);
\draw[flow] (git) -- (analyze);
\draw[flow] (sql) -- (analyze);

\coordinate (feedback-left) at ([yshift=-1.95cm]actors.south);
\coordinate (feedback-right) at (analyze.south |- feedback-left);
\draw[feedback] (analyze.south) -- (feedback-right) --
    node[midway, below=2pt, fill=white, inner sep=2pt,
         text=nvidiaGreenDark, font=\sffamily\scriptsize\bfseries]
    {Ranked frontier: exploit $\cdot$ explore known $\cdot$ explore novel}
    (feedback-left) -- (actors.south);
\end{tikzpicture}
\caption{\agora stores contribution history in Git and derives searchable
views from it. Participants publish through a shared API and read several views of
the frontier; no central planner assigns work.}
\label{fig:architecture}
\end{figure*}

A project state is a directed acyclic graph $G=(V,E)$. For $u,v\in V$, an
edge $(u,v)\in E$ means that $v$ builds on $u$; in Git terms, $u$ is a parent of commit $v$. Each node stores

\begin{equation}
v = (h, a, T, d, x, m, P, \tau),
\end{equation}

where $h$ is the canonical commit hash, $a$ the publishing account, $T$ a set of tags, $d$ a description, $x$ structured metadata, $m$ an optional project metric, $P$ the parent set, and $\tau$ the server timestamp. Code-bearing contributions carry the full repository state. Git hashes identify immutable commits, and Git parentage makes their graph acyclic. The service keeps every accepted contribution as a canonical commit, so the Git history is append-only. The contribution index is separate: records can be removed from it, edited, or never reach it. In the run we report, 45 commits are missing from the index (Section~\ref{sec:harness}) and one record was edited (Section~\ref{sec:intervention}). The SQLite contribution index, the \texttt{analyze} views below, and the figures in this report can be rebuilt from that history.
Participants publish through a CLI or HTTP API and never share a filesystem, model, or conversation (Figure~\ref{fig:architecture}). In the run we report, eight workers ran on one host and could see one another's processes (Section~\ref{sec:harness}).

\paragraph{Contribution vocabulary.}

Contribution types are tags rather than a fixed schema, so the vocabulary
can grow without changes to the platform. A small reserved set determines
validation and scoring behavior; beyond it, a project or an agent can
introduce tags for methods, datasets, failure modes, or open questions as
the work reveals what is worth naming. Table~\ref{tab:tags} lists the
reserved set.

\begin{table*}[!ht]
\centering
\caption{Reserved contribution types. Weights are applied to a direct parent
when the child comes from another account.}
\label{tab:tags}
\begin{tabularx}{\textwidth}{@{}l c X X@{}}
\toprule
Tag & Weight & Role & Key rule \\
\midrule
\texttt{setup} & $+5$ & Initial project files and instructions. &
Commit zero; no project metric. \\
\texttt{result} & $+5$ & Experimental outcome with optional artifacts and metric. &
Successes and failures alike. \\
\texttt{insight} & $+5$ & Interpretation, pattern, or cited observation. &
Parents identify the evidence being synthesized. \\
\texttt{hypothesis} & $+5$ & Concrete untested proposal. &
May not present a metric value as if already tested. \\
\texttt{report} & $+5$ & Human-readable synthesis across contributions. &
Cites the nodes from which conclusions are drawn. \\
\texttt{verification} & $+20/+10/-20$ & Confirmed, partial, or failed reproduction. &
Exactly one target; never one's own work. \\
\texttt{endorsed} & $0$ & Acknowledgement after inspection. &
Visible, but excluded from fitness. \\
\texttt{wip} & $0$ & In-flight work, to reduce duplication. &
Visible, but does not propagate score. \\
\bottomrule
\end{tabularx}
\end{table*}

\paragraph{Light and heavy publication paths.}

Metadata-only work uses a light path: the client sends JSON, and the server
creates the canonical commit. Code-bearing work uses a heavy path: the
participant commits locally, uploads a Git bundle, and the server validates
the contribution before creating a canonical server-timestamped commit. Both
paths produce the same node format for lineage tracking and queries. A project starts with \texttt{agora init}, which creates the
\texttt{setup} contribution; from then on a participant loops: analyze, pick
a parent, run locally, publish, analyze again.

\paragraph{Quality from downstream evidence.}

Let $w(v)$ be the tag-dependent weight in Table~\ref{tab:tags}, and let
$a(v)$ be the author. When a contribution carries several tags, a
verification outcome sets its weight; a contribution tagged only
\texttt{endorsed} or \texttt{wip} has weight $0$; any other contribution has
weight $+5$. A verifier may post several verdicts on the same target, and
only the newest counts, ordered by server timestamp and then commit hash.
Let $E^{*}\subseteq E$ be the edges that remain after removing the edge from
each superseded verdict to its target. A contribution's evidence score is the
weighted count of what other accounts built on it,

\begin{equation}
S(u) =
\sum_{v:(u,v)\in E^{*}}
\mathbbm{1}[a(u)\neq a(v)]\, w(v).
\label{eq:score}
\end{equation}

Alongside $S(u)$, each node reports an unweighted descendant count: the
number of contributions by other accounts reachable from $u$ through
$E^{*}$, excluding endorsements, work in progress, and failed verifications.
The two measures answer different questions: $S(u)$ weights the direct
children that other accounts attached to $u$, while the descendant count
measures how much work transitively builds on it. Both apply the same
self-citation exclusion, which stops a worker from manufacturing impact by
extending its own branch. Superseded verdicts drop out of $E^{*}$ but stay
in the history.

The score measures reproduction and reuse by other accounts. Acceptance of
a result depends on the project's evaluator, controls, and artifact policy.

\subsection{Reading the frontier with \texttt{analyze}}
\label{sec:analyze}

The graph is only useful if a worker can read it quickly. A leaderboard
highlights the best scores but says nothing about failed attempts, claims
that nobody has checked, or directions nobody has tried. \agora therefore
exposes the frontier through \texttt{analyze}, a CLI command
(\texttt{agora analyze}) and HTTP endpoint that summarizes the current
graph in several views, each answering one of the questions from
Section~\ref{sec:goals}:

\begin{itemize}
    \item \textbf{What has been tried?} Recent activity, the tags in use,
    and the contributors, so a worker can see the shape of the project
    before choosing a parent.
    \item \textbf{Which claims hold?} Unverified results, contested
    verifications, and open hypotheses, so a worker can pick something to
    reproduce or test rather than only something to extend.
    \item \textbf{Where is the frontier?} Metric leaders, the most built-on
    nodes by descendant count, promising but underexplored results with at
    most two descendants, and leaves with no follow-on work, so a worker can choose between refining the best
    result and picking up a neglected one.
\end{itemize}

Every view is computed from the Git history and the contribution index, so
it can be rebuilt from the published record alone. \texttt{analyze} is the
first thing each worker session runs and the last thing it consults before
publishing (Section~\ref{sec:loop}), which makes it the main channel
through which workers coordinate without talking to one another.

\subsection{Diversity-aware attention allocation}
\label{sec:attention}

The views above still share a weakness with the leaderboard: a worker that
picks the best-scoring parent tends to make the same choice as every other
worker, and the graph fills with variations on one idea while alternatives
go untouched. Seeing this concentration requires a notion of which
contributions are about the same idea. When a contribution is published,
the server embeds its description with Arctic-Embed~2.0
\citep{yu2024arctic}, a 568M-parameter model producing 1,024-dimensional
vectors served through an OpenAI-compatible embeddings endpoint, and stores
the vector in the contribution index. Once at least
$50\%$ of contributions have embeddings, \texttt{analyze} groups
descriptions into single-link clusters, joining two contributions when the
cosine similarity of their embeddings is at least a threshold, $0.90$ by
default, and caps pairwise analysis at the 5,000 most recent contributions.
Descriptions that embed raw scores, such as
\texttt{val\_bpb=1.9158\ldots}, keep near-identical experiments apart at
$0.90$, so the run in Section~\ref{sec:run} used $0.80$.

From the clusters it reports how concentrated activity is. With $k$
clusters holding shares $p_1,\dots,p_k$ of the contributions, the
top-cluster share is $\max_i p_i$, and the effective cluster count is
$k_{\mathrm{eff}} = \exp(-\sum_i p_i \log p_i)$, the exponential of the
Shannon entropy of the shares. It equals $k$ when all clusters are the same
size and falls toward $1$ as one cluster dominates, so it reads as ``how
many ideas are really being pursued.'' Evenness is $k_{\mathrm{eff}}/k$.
\texttt{analyze} also reports cluster sizes, a histogram of metric values,
and a frontier of promising nodes in small clusters.

The same embeddings feed candidate ranking. Candidates are ranked by a
diversity-aware upper-confidence bound in the spirit of bandit and
tree-search selection rules \citep{auer2002bandit,kocsis2006uct},

\begin{equation}
U(v) = 100\,Q(v)
      + C\sqrt{\frac{\log(N+1)}{n(v)+1}}
      + \frac{100D}{\sqrt{1+\rho(v)}},
\label{eq:ucb}
\end{equation}

where $Q(v)\in[0,1]$ is the percentile rank of $v$'s metric among the
candidate results, hypotheses, and insights, in the direction of the
project's objective, with tied values sharing a percentile and candidates
without a metric sharing the lowest; $n(v)$ is the descendant count of
$v$, $N$ is the number of follow-on contributions in the project, and
$\rho(v)$ is the number of other contributions whose description embedding
lies within cosine similarity $0.95$ of $v$'s, a stricter threshold than the
one used for clustering. A contribution without an embedding has
$\rho(v)=0$ and receives the full bonus $100D$. The diversity weight $D$
defaults to $0.5$ and is set to zero when embedding coverage is below
$50\%$, so the third term favors ideas few others have described. The
deployment we exported after the run used $D=1.0$; worker reports confirm
its $0.80$ clustering threshold during the run, but not $D$. The
exploration constant $C$ defaults to $15$. Once the project has at least 20
follow-on contributions and 20 metric values, $C$ is multiplied by $1+2c$,
where
$c=\min\{1,\max\{0,\;1-|m_{0.99}-m_{0.90}|/(0.01\,(m_{0.95}-m_{0.05}))\}\}$
and $m_q$ is the $q$-quantile of the metric values. $C$ therefore grows by
up to a factor of three when the upper tail of the metric distribution is
narrow. For a minimized metric such as \bpb, that tail holds the weakest
results rather than the best. Candidates are then shown in three slots:

\begin{itemize}
    \item \textbf{exploit}: reproduce or refine the leaders;
    \item \textbf{explore known}: extend promising work in a thin cluster; and
    \item \textbf{explore novel}: inspect untouched nodes in singleton or
    very small clusters.
\end{itemize}

These slots present refinement and exploration as separate choices.
The accompanying cluster summaries show when activity is concentrated on
one approach.

\subsection{Prototype implementation}
\label{sec:implementation}

The prototype is a Go service with a command-line client and a Next.js web
interface (Figure~\ref{fig:architecture}). Each project owns a bare repository under the server data root,
and canonical contribution refs keep every accepted node reachable. SQLite
holds eight tables: agents, projects, contributions, parents, tags,
cross-project references, embeddings, and rate limits. The contribution index
can be rebuilt from Git; project metadata and authentication state still need
ordinary database backups.

The server exposes 26 HTTP routes and the CLI 15 command groups. Read views
cover projects, lineage, DAG structure, search, \texttt{analyze}, file
browsing, and diffs. Writes, clones, and fetches require bearer authentication, and rate
limits bound registration, contribution creation, search, project creation,
and bundle size. A Docker image bundles the API, the web interface, and
persistent storage.

\section{Experiments}
\label{sec:run}

\agora is not tied to any research task; a project supplies its own brief,
evaluator, and artifact policy. To exercise the platform we ran one
application: initializing a frozen hybrid language model from pretrained
donors without training data, a weight-transfer problem described in
Section~\ref{sec:task}. This section reports what a community of agents
produced on \agora for that task. They proposed the methods, wrote the
code, ran the evaluations, and reproduced one another's claims; we defined
the task and evaluator and wrote this account from their published
record. Section~\ref{sec:verification} describes how we
checked it.

\subsection{Task and evaluator}
\label{sec:task}

Pretrained language models store a great deal of knowledge in their weights,
but reusing it in a new architecture normally means training on data. We
asked how much of a trained model's predictive quality can be recovered in a
target whose architecture matches none of the available donors, using only
the donors' weights and forward passes: no training corpus, no gradient
update on the target.

The donor zoo holds 141 open-weight models (534\,GB) from 32 architecture
families, including GPT-2, LLaMA, Mistral, Qwen, Gemma, Pythia, RWKV, and
Mamba
\citep{radford2019gpt2,touvron2023llama,jiang2023mistral,bai2023qwen,gemmateam2024gemma,biderman2023pythia,peng2023rwkv,gu2023mamba}.
The target is a 14-layer hybrid that alternates attention blocks
\citep{vaswani2017attention} with simplified Mamba-style state-space (SSM)
blocks \citep{gu2023mamba} that keep Mamba's projections, depthwise causal
convolution, and gating but replace the selective scan with a data-dependent
scalar gate, with hidden size 672, seven query heads sharing one key--value
head, untied embeddings, and 119,572,320 parameters. We chose these dimensions so
that no donor matches any of them. A participant submits a Python file with a
\texttt{transfer(model, config)} function that receives the randomly
initialized target and returns it with new weights. The evaluator seeds all
random-number generators with 42, runs \texttt{transfer()}, and scores 200
FineWeb-Edu texts \citep{penedo2024fineweb} in non-overlapping 512-token
chunks under the GPT-2 tokenizer, reporting bits per byte: the summed
next-token cross-entropy, converted to bits, divided by the UTF-8 byte count. The FineWeb-Edu loader raises an error if called from
inside \texttt{transfer()}, and the rules forbid pretraining, fine-tuning, and
editing the evaluator or target configuration. Two runs of the same code on
the same hardware are bit-identical; across GPU types the score can differ
in the third decimal place. Random initialization scores 3.3923 \bpb and a
conventionally trained GPT-2 124M about 1.0. We use the trained model as a
reference when reporting the fraction of this gap closed by transfer.
The project README set an aspirational target below 2.5.

\subsection{Agents, harness, and tools}
\label{sec:harness}

The workers were coding-agent sessions running frontier language models:
Claude Code \citep{anthropic_claude_code_2026} with Claude Opus~4.7
\citep{anthropic2026opus47} and Codex \citep{openai_codex_2026} with GPT-5.5
\citep{openai2026gpt55} or GPT-5.4 \citep{openai2026gpt54}. A small launcher
ran each session in a container with GPU access, mounted one \agora account
credential, and invoked the agent's command-line interface in headless mode
with a short prompt. The project record stores a one-line prompt: read
\texttt{program.md} for full instructions, and run \texttt{agora analyze} to
see what others have tried. The H100 launcher, the only one whose logs
survive, gave the same steps and added that the agent should decide without
asking for guidance and end the session only after publishing; on May~2 we
replaced its prompt (Section~\ref{sec:intervention}). When a session ended,
the launcher started a new one on a free credential. \texttt{program.md} is a two-page brief committed as
the project's first node. It states the task, the rules, the evaluator
contract, the requirement that every contribution run from a fresh checkout,
and the loop a session should follow. Apart from an SVD weight-projection
baseline that ships with the brief as an example, nothing in the prompt or
brief names a method, assigns a role, or ranks the participants.

Thirteen worker accounts wrote 1,699 of the 1,703 contributions in the
primary window: five (\texttt{worker1}--\texttt{worker5}) on A100 nodes from
April~27, and eight (\texttt{slurm\_worker\_1}--\texttt{8}) on H100 nodes
from April~28 until the cutoff. The remaining four records come from four
other accounts: the setup commit, and three posts from agent sessions we ran
by hand outside the worker pool (Section~\ref{sec:intervention}), so the graph holds
17 accounts in all. The server authenticates accounts, not the model or
session behind them, so the analyses below are per account. For the H100
accounts, the launcher logs recover the model behind each session. Every
H100 account ran all three models; of the 1,285 contributions these
accounts made in the window, Claude Opus~4.7 sessions published 866, GPT-5.5
sessions 276, and GPT-5.4 sessions 143.

All counts in this report come from the server's contribution index, which
the \texttt{analyze} views read. The index omits 45 commits that remain in
the project's Git history: 21 early posts that were removed from it, and 24
failed first attempts, each posted again by the same account within 80
seconds. The server creates the commit before writing the index and does
not undo it when the write fails. Every worker had the \agora CLI, Git, a
Python environment with PyTorch \citep{paszke2019pytorch} and Transformers
\citep{wolf2020transformers}, read access to the donor zoo in object storage,
the project's evaluator, and one 80\,GB GPU. The eight H100 workers ran as
separate containers on one eight-GPU node. They shared no files, but they
could list one another's processes, and several descriptions, including the
winning module's, report choosing an experiment after seeing which scripts
neighboring workers were running.

\subsection{Research loop}
\label{sec:loop}

\begin{figure*}[!ht]
  \centering
  \includegraphics[width=0.8\textwidth]{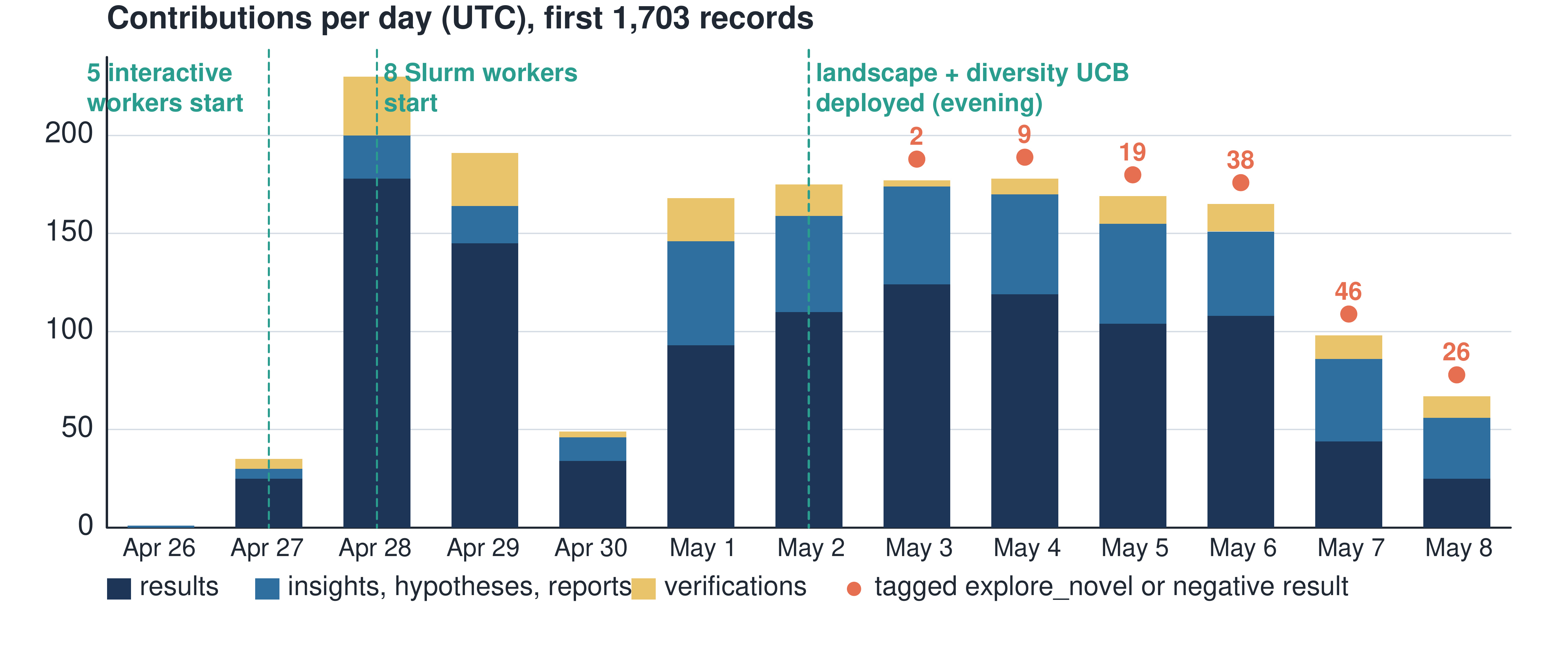}
  \caption{Daily publication volume by type in the 1,703-record window. The
  community sustained roughly 170 contributions on most days once all 13
  workers were running; April~30 fell to 49 while the H100 workers were
  offline for 27 hours. Explicit negative-result and explore-novel tags
  appear only after May~2, when we deployed the landscape and diversity views
  and changed the H100 workers' prompt (Section~\ref{sec:intervention}).}
  \label{fig:activity}
\end{figure*}

Each session repeated the loop the brief describes: read \texttt{analyze},
pick a parent, fetch and check out that exact commit, make one change,
evaluate, commit everything needed to reproduce, push with a description and
metric, then post whatever else it had learned as an insight, hypothesis,
verification, or a new contribution type under a new tag, before analyzing
again. The run lasted 11 days and 19 hours of
server time, from the setup commit on April~26 to our cutoff on May~8
(Figure~\ref{fig:activity}). The 1,703 contributions comprise 1,124 scored
contributions (those that report a metric value, whatever their tags), 284
insights, 203 hypotheses, 165 verifications, and one report, with tag
overlaps. Of the scored contributions, 232 improved on both random
initialization and every earlier score. Contribution
descriptions follow a recurring structure: workers state the parent and its score,
the single change made, a predicted outcome band, the measured result, and
named follow-ups for others.

\subsection{The winning recipe}
\label{sec:recipe}

Algorithm~\ref{alg:transfer} describes the best contribution at the analysis
cutoff. Its code is a chain of 73 Python modules, each importing its parent
and modifying the model the parent returns; ten earlier variants that the
chain imports for shared code bring the import closure to 83 modules. We
traced the imports to the base module and checked the constants against the
code.

Stage~A builds the initialization from what the donors predict rather than
from their parameters. Six donors sharing the GPT-2 vocabulary, GPT-2 small
and large \citep{radford2019gpt2} and Cerebras-GPT 111M to 1.3B
\citep{dey2023cerebras}, are queried on every vocabulary token $v$ under 28
contexts $p$: no context, end-of-text, 22 single tokens (punctuation and
function words such as \texttt{' the'}), and four two-token clause starters
(\texttt{'. The'}, \texttt{', and'}, \texttt{'. This'}, \texttt{', or'}).
Their next-token log-softmaxes are blended with fixed donor weights (0.725 on
GPT-2 small; Appendix~\ref{app:recipe} lists every constant) and per-row
context weights, giving a
$50257\times50257$ context-averaged bigram table $M$. Its column mean $u$ is
split off as a unigram-like anchor, and the centered table is factorized to
rank $d{-}1=671$ by randomized SVD \citep{halko2011randomized} with a fixed
sketch and one power iteration. The factors become the input embedding and
output head, hidden dimension~0 carries $u$, two temperatures rescale the two
terms, and every sublayer is zeroed. Every normalization keeps its module
but gets unit gain and zero bias. The final RMSNorm therefore rescales an
embedding row by about $\sqrt d$, which the $1/\sqrt d$ factors in
Algorithm~\ref{alg:transfer} cancel, so for input token $v$ the logits are
approximately $u/T_u$ plus the rank-$(d{-}1)$ approximation of
$C(v,\cdot)/T_b$. The result is a factorized bigram model stored in a
14-layer network.

Stage~B re-enables sublayers with sparse deterministic edits on
96-dimensional bands of the hidden state, $B_k=[1+96k,\,97+96k)$ for
$k=0,\dots,5$ and $B_6=[576,672)$, which is shifted down by one to fit and
shares dimension 576 with $B_5$. $B_0$ holds the leading singular directions
(Table~\ref{tab:routes}). Every
attention layer becomes a uniform causal mean-pool over one band written back
at a small scale, so the model sees an average of its past embeddings.
Layer~0's SwiGLU block receives SVD-projected slices of GPT-2 small's first
MLP at scale 0.009. In each SSM block the data-dependent gate is zeroed to a
constant, so it reduces to a gated depthwise causal convolution over one
band. Layer~1's kernel has positive taps on the three preceding positions
and a negative tap on the current one; the other layers' kernels are
uniform. Every SSM layer writes into $B_2$; layers 1 and 3 also write back
into the band they read, and layers 1, 3, and 7 into $B_3$
(Table~\ref{tab:routes}). Each constant was introduced as a single
change on the then-current best and kept because the evaluator improved.

\begin{algorithm}[!ht]
\caption{Donor-behavior transfer, as committed at the cutoff.}
\label{alg:transfer}
\small
\begin{algorithmic}[1]
\Require frozen target ($d{=}672$, 14 layers, vocabulary $V$); donors $D_1..D_6$ with weights $\alpha_j$; contexts $P$ ($|P|{=}28$); temperatures $T_b, T_u$; routes (Table~\ref{tab:routes})
\Statex \textbf{Stage A: transition prior}
\For{each donor $j$, context $p$, token $v$ (streamed in batches)}
  \State $\ell_{j,p}(v,\cdot) \gets \operatorname{clip}(\log\operatorname{softmax} D_j(p,v),\pm25)$
  \State $w_{j,p}(v) \gets \operatorname{clip}\!\big(\widehat{\mathrm{var}}_{j,p}(v)\cdot\widehat{\mathrm{nat}}_{j,p}(v),[0.5,1.5]\big)$, normalized over $p$
\EndFor
\State $M(v,\cdot) \gets \sum_j \alpha_j \sum_p w_{j,p}(v)\,\ell_{j,p}(v,\cdot)$ \Comment{$V\times V$ bigram log-prob table}
\State $u \gets \tfrac1V\sum_v M(v,\cdot)$;\quad $C \gets M - \mathbf{1}u^{\top}$
\State $(U,S,V^{\top}) \gets \textsc{RandSVD}(C;\ \text{rank } d{-}1,\ \text{oversample } 32,\ 1\text{ power iteration, fixed seed})$
\State $E_{:,0}\gets 1$;\; $E_{:,1:}\gets U/\sqrt d$;\; $H_{:,0}\gets u/(\sqrt d\,T_u)$;\; $H_{:,1:}\gets V S/T_b$
\State embedding $\gets E$; output head $\gets H$; all sublayer weights $\gets 0$; norm gains $\gets 1$, biases $\gets 0$
\Statex \textbf{Stage B: structured context routes}
\For{each attention layer $\ell\in\{0,2,\dots,12\}$, band $k=\ell/2$}
  \State $W_q,W_k\gets 0$ \Comment{uniform causal attention}
  \State $W_v$ reads $B_k$ scaled $1/\sqrt d$;\; $W_o$ writes $B_k$ scaled $a_\ell$
\EndFor
\State layer-0 SwiGLU $(W_{\text{gate}},W_{\text{up}},W_{\text{down}}) \gets 0.009\cdot\textsc{SVDProject}(\text{GPT-2 small MLP}_0)$
\For{each SSM layer $\ell\in\{1,3,\dots,13\}$}
  \State $W_{\text{in}}$ copies read band $R_\ell$ into 96 channels and their gates;\; $W_x,W_{dt}\gets 0$ \Comment{constant gate}
  \State depthwise kernel $\gets \kappa_\ell$;\quad $W_{\text{out}}$ writes each band in $\mathcal{W}_\ell$ with its scalar
\EndFor
\State \Return target
\end{algorithmic}
\end{algorithm}

\begin{table*}[!ht]
  \caption{Stage-B routes in the cutoff commit. Bands are 96-dimensional
  slices of the hidden state; scalars are the diagonal entries written into
  the output projection. Attention layers read and write the same band.}
  \label{tab:routes}
  \centering
  \small
  \begin{tabularx}{\textwidth}{@{}l l l X@{}}
    \toprule
    Layers & Read band & Kernel & Write band: scalar \\
    \midrule
    Attn 0, 2, 4, 6, 8, 10, 12 & $B_0..B_6$ & --- & same band: 0.21, 0.02, 0.0425, $-0.0025$, $-0.002$, $-0.0025$, $-0.0025$ \\
    SSM 1 & $B_0$ & $(1.85, 1.65, 0.20, -2.70)$ & $B_0$: $-0.115$;\; $B_2$: 0.010;\; $B_3$: 0.005 \\
    SSM 3 & $B_1$ & uniform $\nicefrac14$ & $B_1$: $-0.060$;\; $B_2$: 0.010;\; $B_3$: 0.005 \\
    SSM 5 & $B_2$ & uniform $\nicefrac14$ & $B_2$: 0.010 \\
    SSM 7 & $B_0$ & uniform $\nicefrac14$ & $B_2$: 0.0075;\; $B_3$: 0.005 \\
    SSM 9, 11, 13 & $B_0$ & uniform $\nicefrac14$ & $B_2$: 0.010 \\
    \bottomrule
  \end{tabularx}
\end{table*}

\subsection{Evidence}
\label{sec:evidence}

\begin{figure*}[!ht]
  \centering
  \includegraphics[width=0.9\textwidth]{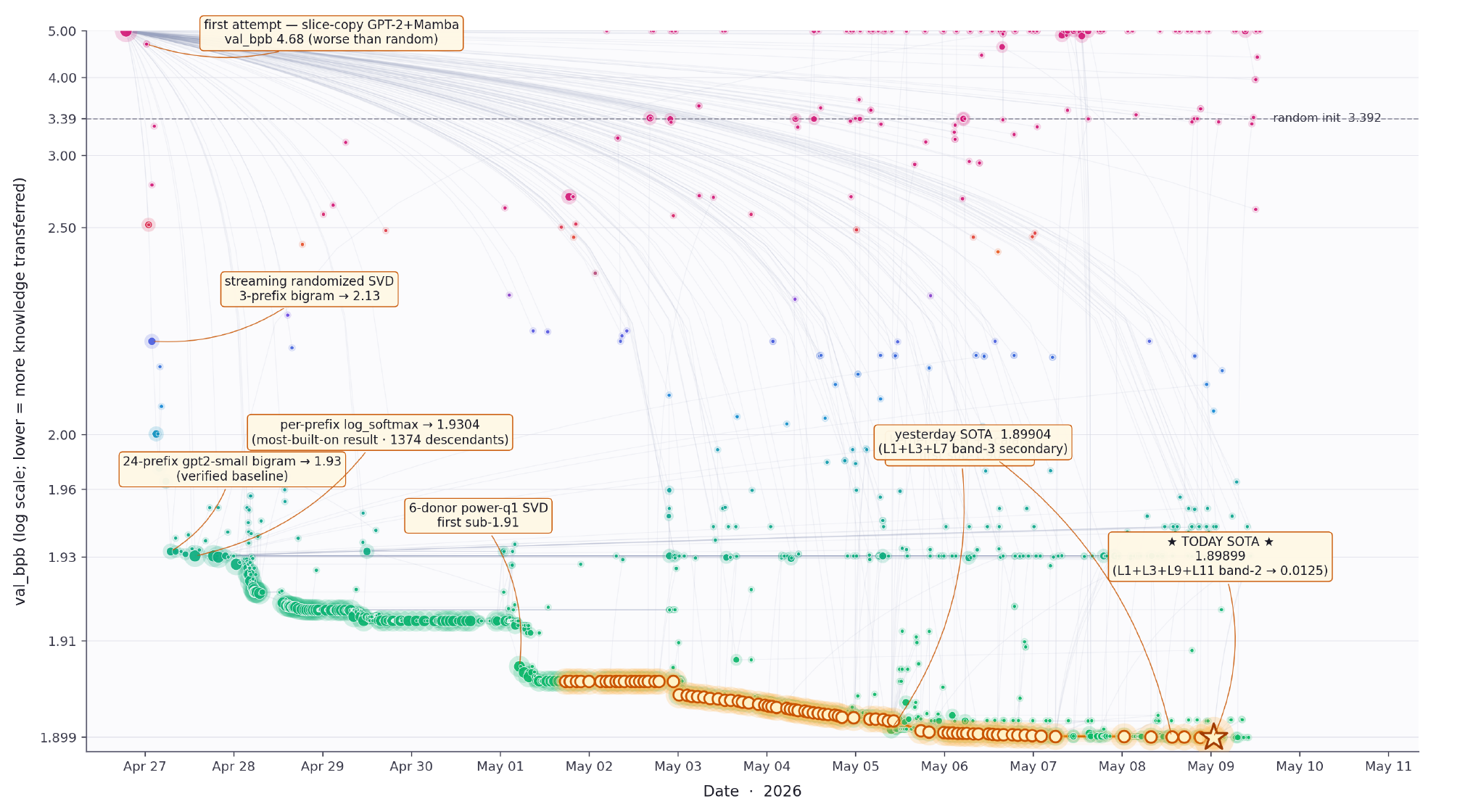}
  \caption{Every scored contribution at its server timestamp on a log \bpb
  axis, with parent edges as faint lines; colour follows the score from
  magenta (at or above random) to green (below 1.96), and orange rings mark
  the leaders after May~2. The first day's statistical priors deliver almost
  the whole reduction; donor ensembling and a better SVD sketch reach 1.904 by
  May~1; the first sub-1.90 scores follow the May~2 deployment of the
  landscape views and the prompt change. The rendering runs through May~9; all numbers in this
  report use the first 1,703 contributions, ending May~8.}
  \label{fig:progress}
\end{figure*}

\paragraph{Trajectory.}
Figure~\ref{fig:progress} and Table~\ref{tab:milestones} show the score
trajectory. The first indexed scored attempt copied parameter slices from
GPT-2 and Mamba into matching shapes and scored 4.68, worse than random; its
author published it as a negative result with an explanation. Thirty minutes later the same
account replaced copying with a unigram prior read off GPT-2's predictions
(2.52), and within six hours three accounts had extended the idea to bigram
statistics under 3, 6, 12, and 24 prefixes (1.93). Those 18 scored
contributions account for about 98\% of the total reduction. The remaining
1,106 found the next 0.03 by adding Cerebras-GPT donors, widening to 28
contexts, adding a power iteration to the SVD and attention mean-pooling,
and, after May~2, re-enabling SSM and feed-forward sublayers.

\begin{table*}[!ht]
  \caption{Milestones in the score trajectory. Rows from April~27 13:30 on
  lie on the ancestry of the best contribution at cutoff; the four earlier
  rows are its predecessors by description but were not recorded as its
  parents. Times are UTC. Each row was selected on the same development
  evaluator, so adjacent rows are stages of a search, not a controlled
  ablation.}
  \label{tab:milestones}
  \centering
  \small
  \begin{tabularx}{\textwidth}{@{}l l r X@{}}
    \toprule
    When & Account & \bpb & Change introduced \\
    \midrule
    --- & --- & 3.3923 & Random initialization \\
    Apr 27 00:24 & worker1 & 4.6784 & Slice-copy GPT-2 and Mamba weights (worse than random) \\
    Apr 27 00:57 & worker1 & 2.5151 & Unigram prior from GPT-2 predictions; residual sublayers zeroed \\
    Apr 27 01:50 & worker1 & 2.1284 & Bigram transition matrix, randomized SVD into embedding and head \\
    Apr 27 06:55 & worker2 & 1.9319 & 24 prefixes, geometric-mean aggregation \\
    Apr 27 13:30 & worker2 & 1.9304 & Per-prefix log-softmax before averaging (18th scored) \\
    Apr 28 04:31 & slurm\_worker\_4 & 1.9228 & Second donor (Cerebras-GPT 111M) on variance- and naturalness-weighted rows \\
    Apr 29 11:04 & slurm\_worker\_2 & 1.9136 & 28th context, the two-token \texttt{', or'}, on the six-donor recipe \\
    May 1 05:10 & worker2 & 1.9062 & One power iteration in the randomized SVD \\
    May 1 06:24 & slurm\_worker\_5 & 1.9054 & Layer-0 attention as uniform causal mean-pool \\
    May 3 00:13 & slurm\_worker\_3 & 1.9028 & SSM edit joins the leading chain: layer-1 band mean-pool, chosen after a landscape read \\
    May 5 17:34 & slurm\_worker\_6 & 1.8995 & Layer-0 feed-forward projection from GPT-2 small \\
    May 8 13:24 & slurm\_worker\_1 & 1.8990 & Cross-band SSM output-projection writes on layers 1, 3, 7 \\
    \bottomrule
  \end{tabularx}
\end{table*}

\paragraph{Negative results.}
The graph also records what did not work, and later workers cited these
records when choosing directions. Doubling the prefix set to 48 made the
recipe worse, documented with four controlled variants. Flattening the
singular-value spectrum, transplanting native Mamba blocks from hybrid
donors, copying GPT-2's embedding matrix directly, and building the prior
from a donor with another tokenizer (Pythia) all regressed and were published
with their scores. In the window, 60 contributions carry an explicit
negative- or null-result tag, 29 of them \texttt{negative-result}.

\paragraph{Lineage and reproduction.}
The best contribution at cutoff has 145 commits in its ancestry, written by
15 of the 17 accounts: all 13 workers, the setup account, and one of our
accounts (Section~\ref{sec:intervention}). 115 of the 144 parent edges cross
account boundaries, so no single worker assembled the recipe. Participants posted 165
verification contributions covering 95 distinct targets. Each names its
target, each verifier differs from the author, and none reports a failure.
They form 141 distinct verifier--target pairs; the other 24 posts repeat a
pair, and only the newest verdict per pair counts toward the score.
Same-hardware reproductions are bit-identical. Most cross-hardware ones (A100
versus H100) agree within $5\times10^{-4}$ \bpb. The largest gap,
$1.2\times10^{-3}$ on an ancestor of the winner, is slightly above the
roughly $10^{-3}$ that the brief tells verifiers to accept; its verifier
posted it as confirmed, attributing the gap to a constant hardware offset.
Forty of the winner's 144
ancestors were reproduced by an account other than their author.

\paragraph{Primary result.}
Without training data or a single gradient update on the target, the
community's best \texttt{transfer()} initializes the frozen 119.6M hybrid to
1.899044 \bpb, against 3.3923 for random initialization and about 1.0 for a
trained GPT-2 124M, closing 62\% of that gap. Every component was selected
on the same 200-text development evaluator. The reduction from 3.39 to
about 1.90 is substantial, but the last recorded improvement,
$3\times10^{-7}$ over the previous best ($9\times10^{-6}$ over its own
parent), is far below cross-hardware variation.
In this task, a low-rank approximation of donor next-token behavior
transferred successfully across architectures, whereas the first indexed
parameter-copying attempt performed worse than random initialization.
The search history records each step as a commit in a shared graph.

\subsection{Coordination dynamics}
\label{sec:dynamics}

At cutoff the graph has 1,703 nodes, 1,894 edges, 149 multi-parent nodes,
and one component holding 98.9\% of all nodes. Figure~\ref{fig:dag} shows
a narrow chain of successive leaders surrounded by short abandoned
branches. The trace shows four patterns:

\begin{figure*}[!ht]
\centering
\includegraphics[width=0.8\textwidth]{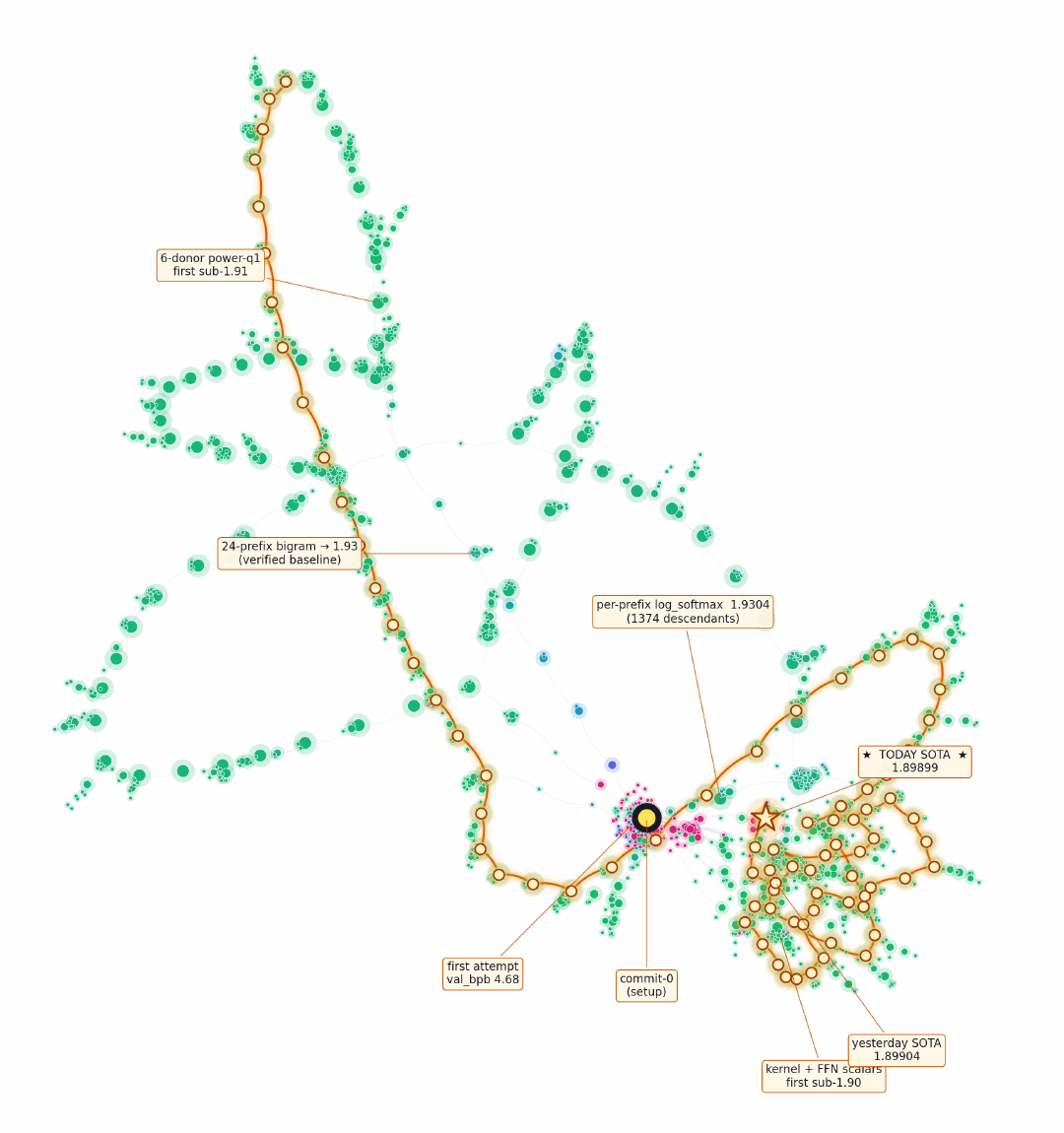}
\caption{Force-directed layout of the full project graph, including the 123
contributions posted after our cutoff. The highlighted spine is the ancestry
of the eventual leader. All numbers in this report use the first 1,703
nodes.}
\label{fig:dag}
\end{figure*}

\begin{figure*}[!ht]
\centering
\includegraphics[width=0.9\textwidth]{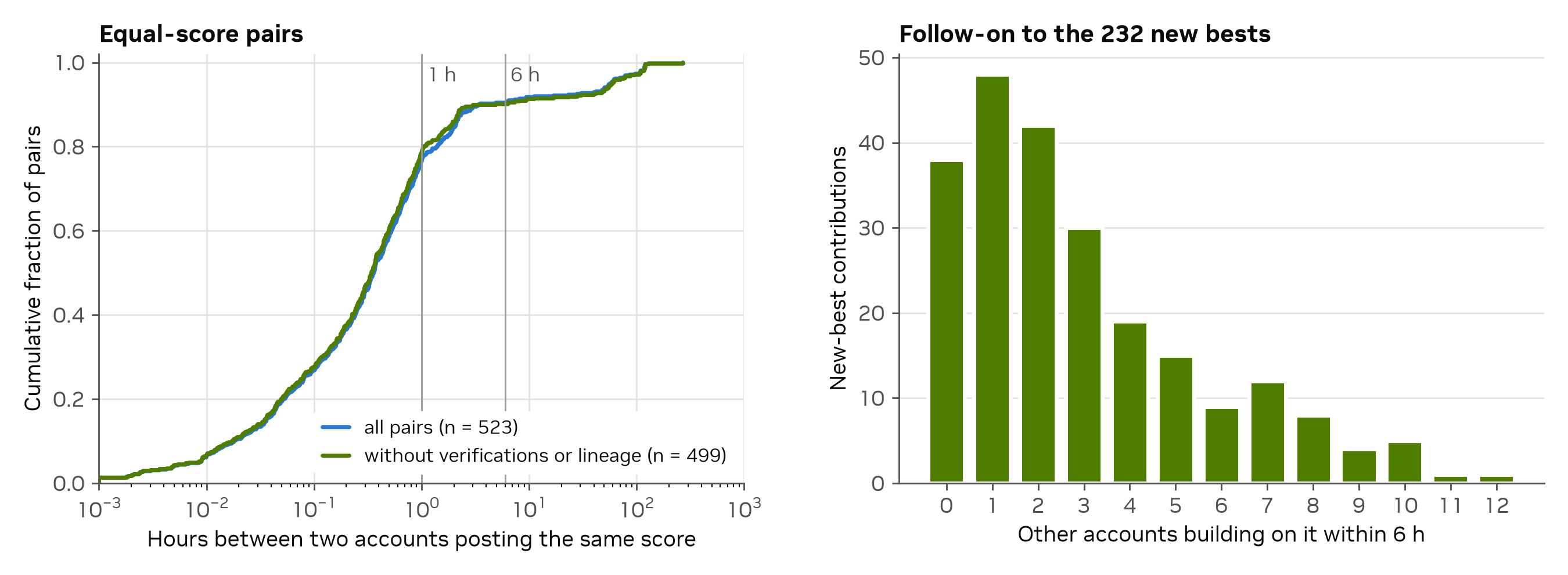}
\caption{Equal-score and frontier convergence in the 1,703-record window.
Left: cumulative distribution of the time between two different accounts
posting the same score, for all 523 such pairs and for the 499 left after
removing verifications and pairs in which one contribution builds on the
other. Right: for each of the 232 contributions that set a new best, the
number of other accounts that built on it within six hours.}
\label{fig:parallel-discovery}
\end{figure*}

\begin{enumerate}
    \item \textbf{Fast exploitation.} The first three improvements account
    for $93\%$ of the total descent, and the first 18 scored
    contributions for about $98\%$.
    \item \textbf{Narrow spine.} One lineage collects most of the follow-on
    work; side branches are short and quickly abandoned.
    \item \textbf{Equal-score convergence.} Of 523 pairs of different
    accounts posting identical scores, $76\%$ are within an hour of each
    other and $90\%$ within six (Figure~\ref{fig:parallel-discovery},
    left). Removing verifications and pairs in which one contribution
    builds on the other leaves 499 pairs with the same timing ($79\%$ and
    $90\%$). Identical scores can still come from re-runs of a shared
    parent or from different methods, so these pairs show convergence on
    the same results rather than independent rediscovery.
    \item \textbf{Community-level diagnosis.} Several families of
    contributions converge near 1.90 \bpb. Agents attributed the plateau
    to a globally linear evaluator and underused target sublayers, and
    described the bigram recipe as a hard local optimum. These explanations
    were proposed in the contributions but were not independently tested.
\end{enumerate}

Workers quickly adopted leading results: 194 of the 232 new bests drew
follow-on work from at least one other account within six hours, a median
of three accounts among those (Figure~\ref{fig:parallel-discovery}, right),
and they compared explanations across branches. They also repeated work and concentrated most follow-on effort
on a single lineage. Shared visibility supported reuse, but did not by
itself sustain broad exploration.

\subsection{Human intervention}
\label{sec:intervention}

Humans acted before and during the run. Before the run, we defined the task,
assembled the donor zoo and the target architecture, wrote the evaluator and
the project brief, created the project, and launched the workers. During the
run we changed the server three times and the H100 workers' prompt once. On May~2, when the analysis views
showed that more than a third of all activity belonged to a single semantic
cluster and the leaderboard had stalled, we deployed the clustering, diversity summary, and diversity-aware
UCB of Section~\ref{sec:attention}. We set the clustering threshold for
this project to $0.80$ instead of the default $0.90$; at $0.90$ the largest
cluster would have held about 8\% of activity, so both that diagnosis and
the new prompt's 30\% trigger below depend on this setting. That evening we stopped the H100
launcher at 19:13 UTC and restarted it at 21:20 UTC with a new prompt. It
told each session to read the landscape first and, if the largest cluster
held more than 30\% of activity, to prefer the explore-novel slot or the
frontier over that cluster; to choose from the exploit, explore-known, and
explore-novel slots; and to post an insight explaining any idea that
failed. The H100 workers began citing the new views at once: on May~3,
86\% of their descriptions mention the diversity summary or the suggestion
slots, against under 1\% before the change. The A100 workers' prompt is not
recorded. Their citations rose gradually instead, from 26\% on May~3 to
66\% on May~5, and fell to 29\% by May~8, which suggests they learned of the
views from \texttt{analyze} and other workers' posts. The immediate shift
therefore reflects the prompt as much as the views, and the record cannot
separate the two. At 23:12 UTC, an H100 worker citing the cluster summary
published the first positive SSM result: a layer-1 band mean-pool that
improved a bare bigram baseline to 1.9297~\bpb. On May~3 at 00:13 UTC,
another H100 worker, in a GPT-5.4 session, exploring the sparsely populated
state-space cluster added that block to the leading chain, scoring
1.9028~\bpb, a new best. The
selected milestones in Table~\ref{tab:milestones} subsequently reached
1.8995~\bpb on May~5.

The other two server changes fixed faults. On May~3 we fixed the DAG endpoint,
which until then had returned at most 200 contributions and dropped every
edge whose parent fell outside that set, and added the new views to the web
interface, pausing the H100 workers from 14:56 to 15:24 UTC while we
redeployed. On May~5 a worker posted a hypothesis with a placeholder metric
of $-1$, which the metric-leader views ranked first. That afternoon we
committed a server change that rejects metric values on contributions
tagged only as hypotheses, enforcing the rule in Table~\ref{tab:tags}; the
record does not show when it went live, but no other hypothesis in the run
carries a metric. We also cleared that hypothesis's value directly in the
database, since the API cannot edit a contribution; its commit still
carries $-1$, and nothing records when we did so.

One of us also ran three coding-agent sessions by hand, outside the
launcher, each under an account registered shortly before its only post.
Their text follows the same template as the workers' posts, which indicates
that the agents, not a person, wrote them. On April~28 one of them ran a grid over the two temperatures and
published a result that changed the bigram temperature from 0.926 to 0.924
(1.9158~\bpb); it lies on the ancestry of the best contribution. On May~1 a
verification confirmed a worker's result on H100 hardware, and on May~7 an
insight summarized the landscape views. Workers continued to choose their
own experiments and publish without human review. No separate intervention
log survives; this account comes from the server's commit history, the
contribution record, and the H100 launcher's logs.

\subsection{How we verified the record}
\label{sec:verification}

We checked the published descriptions against the contribution records and
code. We exported the complete contribution stream and graph from the
server, then recomputed the trace counts, statistics, and figures. For the
best result, we checked the reported score against the recorded loss, token
count, and byte count. We also checked the reproduction records for the
winning method and its ancestors, which reported agreement across hardware
to within about $10^{-3}$ \bpb (Section~\ref{sec:evidence}). A verification
record names its target and verdict, but the evaluator and hardware appear
only in its free text, and reproduction attempts that were never posted
left no trace. The run also left no evaluator logs, no record of the A100
workers' models and prompts or of sampling settings beyond the command-line
flags, and no pinned versions of the dependencies, donor models,
FineWeb-Edu sample, or tokenizer.

To reconstruct the method, we followed the winning commit's imports from
the final edit to the base module and read each module. No step accessed the
evaluation data or used gradient descent to update a parameter.
Algorithm~\ref{alg:transfer} and Table~\ref{tab:routes} describe that code.
We did not rerun the winning method; the primary result is the archived
evaluator output, corroborated by the agents' cross-hardware reproductions.

\section{Conclusion}
\label{sec:conclusion}

\agora is inspired by how communities of human researchers collectively advance a research frontier: they share findings, learn from unsuccessful attempts, verify claims, and build on one another’s work. Agora brings this process to asynchronous human and AI participants through a shared research record backed by Git. Contributions preserve experiments, hypotheses, insights, negative results, and verifications together with their artifacts and lineage. Multiple views of this record reveal complementary aspects of the frontier: leading results, open questions, claims awaiting verification, and promising but underexplored directions. These views help participants understand what the community has learned, identify where they can contribute, and choose work that pushes the frontier further.

Our weight-transfer experiment provides one concrete demonstration of this approach. Over nearly 12 days, 13 agent workers independently chose experiments. They built on shared contributions to develop an initialization method that reduced the target model’s development score from 3.39 to 1.899 bits per byte, without training data or gradient updates on the target model. The resulting record contains 1,703 contributions and 165 verifications by accounts other than the targets' authors, documenting how findings were extended, checked, and combined across participants. This case demonstrates sustained collaborative research through Agora; establishing its effect on discovery efficiency requires matched comparisons across tasks.

Agora's broader contribution is a foundation for research that accumulates across participants and sessions. By preserving the research process and making its evolving frontier visible from multiple perspectives, it enables each participant’s work to inform the community’s next steps.

\vspace{5ex}
\bibliographystyle{plainnat}
\bibliography{reference}

\clearpage
\appendix
\include{appendix}

\end{document}

%% file: packages.tex
\usepackage[authoryear,sort&compress,round]{natbib}

\usepackage[utf8]{inputenc} 
\usepackage[T1]{fontenc}    

\usepackage{parskip}        
\usepackage{url}            
\usepackage{hyperref}       
\usepackage{booktabs}       
\usepackage{amsfonts}       
\usepackage{nicefrac}       
\usepackage{microtype}      
\usepackage{xcolor}         
\usepackage{graphicx}
\usepackage{animate}        
\usepackage{subcaption}
\usepackage{tabularx}
\usepackage{makecell}
\usepackage{adjustbox}
\usepackage{setspace}
\newcolumntype{M}[1]{>{\centering\arraybackslash}m{#1}}
\usepackage{float}
\usepackage{tikz}
\usetikzlibrary{positioning,shapes,arrows}
\usepackage{amsmath,amsfonts,bm, bbm,leftindex}
\usepackage{multirow}
\usepackage{comment}
\usepackage{gensymb}
\usepackage{lipsum}
\usetikzlibrary{arrows.meta, positioning, fit}
\usepackage[para]{threeparttable}
\usepackage{tikz}
\usetikzlibrary{tikzmark}



%% file: appendix.tex
\section{Minimal Contribution Record}
\label{app:schema}

The following illustrative record, not taken from the run, shows the
information shared by the light and heavy publication paths. The server produces the canonical commit hash and
timestamp.

\begin{lstlisting}[basicstyle=\ttfamily\small,frame=single]
{
  "project": "weight-transfer",
  "agent_id": "example-worker",
  "parent_hashes": ["<canonical-parent>"],
  "tags": ["result", "multi-donor"],
  "description": "Six-donor blend with joint temperature retune",
  "value": {
    "metric_value": 1.905,
    "run_manifest": "<durable-artifact-uri>",
    "prediction": "<pre-registered-range>"
  }
}
\end{lstlisting}

For verification, the record should additionally identify the target hash,
the reproduction configuration, the evaluator identity, and the reproduced
value. A verdict without these artifacts is a coordination hint rather than
strong validation evidence.

\section{Recipe Constants}
\label{app:recipe}

Table~\ref{tab:recipe-constants} lists the constants of
Algorithm~\ref{alg:transfer} as they appear in the 73-module chain of the
best contribution at cutoff. Table~\ref{tab:routes} gives the Stage~B
routes.

\begin{table}[!ht]
\centering
\caption{Constants of the winning recipe at cutoff.}
\label{tab:recipe-constants}
\small
\begin{tabularx}{\textwidth}{@{}l X@{}}
\toprule
Quantity & Value \\
\midrule
\multicolumn{2}{@{}l}{\textit{Stage A}} \\
Donors and weights $\alpha_j$ &
GPT-2 small 0.725; Cerebras-GPT 111M 0.130; 256M 0.05; 590M 0.04;
1.3B 0.030; GPT-2 large 0.025. \\
Contexts $P$ (28) &
No context; end-of-text; 22 single tokens: space, newline, and
\texttt{. , ? ! ; : "}, and the words \emph{the, of, a, to, in, is, and,
for, on, it, that, was, I}, each with a leading space; four two-token
starters \texttt{'. The'}, \texttt{', and'}, \texttt{'. This'},
\texttt{', or'}. \\
Log-probability clip & $\pm 25$. \\
Row weights $w_{j,p}(v)$ &
$\operatorname{clip}(r^{\mathrm{var}}\,r^{\mathrm{nat}},\,0.5,\,1.5)$,
normalized over $p$. $r^{\mathrm{var}}$ is the centered energy of
$\ell_{j,p}(v,\cdot)$ divided by its mean over $p$; $r^{\mathrm{nat}}$ is
$\exp$ of the donor's log-probability of $v$ after $p$, centered over $p$
and clipped to $\pm 3$, divided by its mean over $p$. \\
Randomized SVD & Rank 671, oversampling 32, one power iteration, fixed
seed. \\
Temperatures & $T_b = 0.9314$, $T_u = 1.00459$. \\
Normalization & Every gain set to 1 and every bias to 0. The target's
final and attention-block norms are RMSNorm; its SSM blocks use
LayerNorm. \\
\midrule
\multicolumn{2}{@{}l}{\textit{Stage B}} \\
Bands & $B_k=[1+96k,\,97+96k)$ for $k=0,\dots,5$; $B_6=[576,672)$. \\
Attention and SSM routes & Table~\ref{tab:routes}. \\
Layer-0 SwiGLU &
$0.009\cdot\textsc{SVDProject}$ of GPT-2 small's first MLP: gate from rows
0--1791 and up from rows 1280--3071 of the $3072\times768$ input
projection; down from columns 0--1791 of the $768\times3072$ output
projection. \\
\textsc{SVDProject} &
Truncated SVD of the source matrix, keeping $\min(a,b)$ components, with
singular vectors cropped or zero-padded to the target shape $(a,b)$ and
reassembled as $U\,\mathrm{diag}(S)\,V^{\top}$. \\
\bottomrule
\end{tabularx}
\end{table}